\def\year{2019}\relax
\documentclass[letterpaper]{article}
\usepackage{aaai19}
\usepackage{times}
\usepackage{helvet}
\usepackage{courier}
\usepackage{url}
\usepackage{graphicx}
\usepackage{multirow}
\usepackage{algorithm}
\usepackage{algorithmic}
\usepackage{amsmath}
\usepackage{subfigure}

\begin{document}
%
\title{RobustSTL: A Robust Seasonal-Trend Decomposition Algorithm\\ for Long Time Series}
\author{Qingsong Wen, Jingkun Gao, Xiaomin Song, Liang Sun, Huan Xu, Shenghuo Zhu \\
Machine Intelligence Technology, Alibaba Group\\
Bellevue, Washington 98004, USA\\
\{qingsong.wen, jingkun.g, xiaomin.song, liang.sun, huan.xu, shenghuo.zhu\}@alibaba-inc.com\\
}

\maketitle
\begin{abstract}
\begin{quote}
Decomposing complex time series into trend, seasonality, and remainder components is an important task to facilitate time series anomaly detection and forecasting. Although numerous methods have been proposed, there are still many time series characteristics exhibiting in real-world data which are not addressed properly, including 1) ability to handle seasonality fluctuation and shift, and abrupt change in trend and reminder; 2) robustness on data with anomalies; 3) applicability on time series with long seasonality period. In the paper, we propose a novel and generic time series decomposition algorithm to address these challenges. Specifically, we extract the trend component robustly by solving a regression problem using the least absolute deviations loss with sparse regularization. Based on the extracted trend, we apply the the non-local seasonal filtering to extract the seasonality component. This process is repeated until accurate decomposition is obtained. Experiments on different synthetic and real-world time series datasets demonstrate that our method outperforms existing solutions.



\end{quote}
\end{abstract}

\section{Introduction}

With the rapid growth of the Internet of Things (IoT) network and many other connected data sources, there is an enormous increase of time series data. Compared with traditional time series data, it comes in big volume and typically has a long seasonality period. One of the fundamental problems in managing and utilizing these time series data is the seasonal-trend decomposition. A good seasonal-trend decomposition can reveal the underlying insights of a time series, and can be useful in further analysis such as anomaly detection and forecasting~\cite{twitter:anomaly:detection,Laptev:2015:GS,forecast:package,wu2016unfolding}. For example, in anomaly detection a local anomaly could be a spike during an idle period. Without seasonal-trend decomposition, it would be missed as its value is still much lower than the unusually high values during a busy period. In addition, different types of anomalies correspond to different patterns in different components after decomposition. Specifically, the spike \& dip anomalies correspond to abrupt change of remainder, and the change of mean anomaly corresponds to abrupt change of trend. Similarly, revealing the trend of time series can help to build more robust forecasting models.


As the seasonal adjustment is a crucial step for time series where seasonal variation is observed, many seasonal-trend decomposition methods have been proposed. The most classical and widely used decomposition method is the STL (Seasonal-Trend decomposition using Loess)~\cite{cleveland1990stl}. STL estimates the trend and the seasonality in an iterative way. However, STL still suffers from less flexibility when seasonality period is long and high noises are observed. In practice, it often fails to extract the seasonality component accurately when seasonality shift and fluctuation exist. Another direction of decomposition is X-13-ARIMA-SEATS~\cite{X13ARIMA} and its variants such as X-11-ARIMA, X-12-ARIMA~\cite{X12ARIMA}, which are popular in statistics and economics. These methods incorporate more features such as calendar effects, external regressors, and ARIMA extensions to make them more applicable and robust in real-world applications, especially in economics. However, these algorithms can only scale to small or medium size data. Specifically, they can handle monthly and quarterly data, which limits their usage in many areas where long seasonality period is observed. Recently, more decomposition algorithms have been proposed~\cite{dokumentov2015str,rob:tbats,verbesselt2010detecting}. For example, in \cite{dokumentov2015str}, a two-dimensional representation of the seasonality component is utilized to learn the slowly changing seasonality component. Unfortunately, it cannot scale to time series with long seasonality period due to the cost to learn the huge two-dimensional structure. 




Although numerous methods have been proposed, there are still many time series characteristics exhibiting in real-world data which are not addressed properly. First of all, seasonality fluctuation and shift are quite common in real-world time series data. We take a time series data whose seasonality period is one day as an example. The seasonality component at 1:00 pm today may correspond to 12:30 pm yesterday, or 1:30 pm the day before yesterday. Secondly, most algorithms cannot handle the abrupt change of trend and remainder, which is crucial in anomaly detection for time series data where the abrupt change of trend corresponds to the change of mean anomaly and the abrupt change of residual corresponds to the spike \& dip anomaly. Thirdly, most methods are not applicable to time series with long seasonality period, and some of them can only handle quarterly or monthly data. Let the length of seasonality period be $T$, we need to estimate $T-1$ parameters to extract the seasonality component. However, in many IoT anomaly detection applications, the typical seasonality period is one day. If a data point is collected every one minute, then $T=1440$, which is not solvable by many existing methods.


In this paper, we propose a robust and generic seasonal-trend decomposition method. Compared with existing algorithms, our method can extract seasonality from data with long seasonality period and high noises accurately and effectively. In particular, it allows fractional, flexible, and shifted seasonality component over time. Abrupt changes in trend and remainder can also be handled properly. Specifically, we extract the trend component robustly by solving a regression problem using the least absolute deviations (LAD) loss~\cite{lad:lasso} with sparse regularizations. After the trend is extracted, we apply the non-local seasonal filtering to extract the seasonality component robustly. This process is iterated until accurate estimates of trend and seasonality components are extracted. As a result, the proposed seasonal-trend decomposition method is an ideal tool to extract insights from time series data for the purpose of anomaly detection. To validate the performance of our method, we compare our method with other state-of-the-art seasonal-trend decomposition algorithms on both synthetic and real-world time series data. Our experimental results show that our method can successfully extract different components on various complicated datasets while other algorithms fail.

Note that time series decomposition approaches can be either additive or multiplicative. In this paper we focus on the additive decomposition, and the multiplicative decomposition can be obtained similarly. 
The remainder of this paper is organized as follows: Section 2 briefly introduces the related work of seasonal-trend decomposition; In Section 3 we discuss our proposed seasonal-trend decomposition method in detail; the empirical studies are investigated in comparison with several state-of-the-art algorithms in Section 4; and we conclude the paper in Section 5. 

\section{Related Work}

As we discussed in Section 1, the ``classical" method ~\cite{macaulay1931smoothing} continues to evolve from X-11, X-11-ARIMA to X-13-ARIMA-SEATS and improve their capabilities to handle seasonality with better robustness. A recent algorithm TBATS (Trigonometric Exponential Smoothing State Space model with Box-Cox transformation, ARMA errors, Trend and Seasonal Components) ~\cite{rob:tbats} is introduced to handle complex, non-integer seasonality. However, they all can be described by the state space model with a lot of hidden parameters when the period is long. They are suitable to process time series with short seasonality period (e.g., tens of points) but suffer from the high computational cost for time series with long seasonality period and they cannot handle slowly changing seasonality.


The Hodrick-Prescott filter ~\cite{hodrick1997postwar}, which is similar to Ridge regression, was introduced to decompose slow-changing trend and fast-changing residual. While the formulation is simple and the computational cost is low, it cannot decompose trend and long-period seasonality. And it only regularizes the second derivative of the fitted curve for smoothness. This makes it prone to spike and dip and cannot catch up with the abrupt trend changes. 

Based on the Hodrick-Prescott filter, STR (Seasonal-Trend decomposition procedure based on Regression)~\cite{dokumentov2015str} which explores the joint extraction of trend, seasonality and residual without iteration is proposed recently. STR is flexible to seasonal shift, and it can not only deal with multiple seasonalities but also provide confidence intervals for the predicted components. To get a better tolerance to spikes and dips, robust STR using $\ell_1$-norm regularization is proposed. The robust STR works well with outliers. But still, as the authors mentioned in the conclusion, STR and robust STR only regularize the second difference of fitted trend curve for smoothness so it cannot follow abrupt change on the trend. 


The Singular Spectrum Analysis (SSA) ~\cite{danilov1997caterpillar,golyandina2001analysis} is a model-free approach that performs well on short time series. It transforms a time series into a group of sliding arrays, folds them into a matrix and then performs SVD decomposition on this matrix. After that, it picks the major components to reconstruct the time series components. It is very similar to principal component analysis, but its strong assumption makes it not applicable on some real-world datasets.

As a summary, when comparing different time series decomposition methods, we usually consider their ability in the following aspects: outlier robustness, seasonality shift, long period in seasonality, and abrupt trend change. The complete comparison of different algorithms is shown in Table ~\ref{tab:a}.

\begin{table}[htb]
\caption{Comparison of different time series decomposition algorithms (Y: Yes / N: No)}\label{tab:a}
\resizebox{\columnwidth}{!}{%
\begin{tabular}{l|llllll}
\hline
Algorithm & \begin{tabular}[c]{@{}l@{}} Outlier\\ Robustness\end{tabular} & \begin{tabular}[c]{@{}l@{}} Seasonality\\ Shift\end{tabular} & \begin{tabular}[c]{@{}l@{}}Long\\ Period\end{tabular} & \begin{tabular}[c]{@{}l@{}}Abrupt \\ Trend Change \end{tabular} \\ \hline
Classical                            & N                                                                                                                 & N                                                          & N                                                    & N                                                                 \\
ARIMA/SEATS                         & N                                                                                                               & N                                                          & N                                                    & N                                                                 \\
STL                                   & N                                                                                                                    & N                                                          & Y                                                    & N                                                                \\
TBATS                                 & N                                                                                                                 & N                                                         & N                                                    & Y                                                                 \\
STR                                   & Y                                                                                                                 & Y                                                         & N                                                    & N                                                                 \\
SSA                                   & N                                                                                                                  & N                                                         & N                                                    & N                                                                 \\
Our RobustSTL                                & Y                                                                                                                & Y                                                         & Y                                                   & Y                                                                \\ \hline
\end{tabular}%
}
\end{table}


\section{Robust STL Decomposition}\label{section:method}
\subsection{Model Overview} \label{subsection:sysmodel}

Similar to STL, we consider the following time series model with trend and seasonality
\begin{equation}\label{eq:model} 
y_t = \tau_t + s_t + r_t, \quad t = 1,2,...N
\end{equation}
where $y_t$ denotes the observation at time $t$, $\tau_t$ is the trend in time series, $s_t$ is the seasonal signal with period $T$, and the $r_t$ denotes the remainder signal. In seasonal-trend decomposition, seasonality typically describes periodic patterns which fluctuates near a baseline, and trend describes the continuous increase or decrease. Thus, usually it is assumed that the seasonal component $s_t$ has a repeated pattern which changes slowly or even stays constant over time, whereas the trend component $\tau_t$ is considered to change faster than the seasonal component~\cite{dokumentov2015str}. Also note that in this decomposition we assume the remainder $r_t$ contains all signals other than trend and seasonality. Thus, in most cases it contains more than white noise as usually assumed. As one of our interests is anomaly detection, we assume $r_t$ can be further decomposed into two terms: 
\begin{equation}\label{eq:r_t} 
r_t = a_t + n_t,
\end{equation}
where $a_t$ denotes spike or dip, and $n_t$ denotes the white noise.


In the following discussion, we present our RobustSTL algorithm to decompose time series which takes the aforementioned challenges into account. The RobustSTL algorithm can be divided into four steps: 1) Denoise time series by applying bilateral filtering~\cite{bilateral:book}; 2) Extract trend robustly by solving a LAD regression with sparse regularizations; 3) Calculate the seasonality component by applying a non-local seasonal filtering to overcome seasonality fluctuation and shift; 4) Adjust extracted components. These steps are repeated until convergence. 

\subsection{Noise Removal}\label{subsection:denoise} 

In real-world applications when time series are collected, the observations may be contaminated by various types of errors or noises. In order to extract trend and seasonality components robustly from the raw data, noise removal is indispensable. Commonly used denoising techniques include low-pass filtering~\cite{baxter1999measuring,christiano2003band}, moving/median average~\cite{osborn1995moving,wen1999simple}, and Gaussian filter~\cite{blinchikoff2001filtering}. Unfortunately, those filtering/smoothing techniques destruct some underlying structure in $\tau_t$ and $s_t$ in noise removal. For example, Gaussian filter destructs the abrupt change of $\tau_t$, which may lead to a missed detection in anomaly detection. 

Here we adopt bilateral filtering~\cite{bilateral:book} to remove noise, which is an edge-preserving filter in image processing. The basic idea of bilateral filtering is to use neighbors with similar values to smooth the time series $\{y_t\}_{t=1}^N$. When applied to time series data, the abrupt change of trend $\tau_t$, and spike \& dip in $a_t$ can be fully preserved.

Formally, we use $\{y_t'\}_{t=1}^N$ to denote the filtered time series after applying bilateral filtering: 
\begin{equation}\label{eq:denoising} 
y_t' = \sum_{j \in J} w_j^t y_j, \quad J = {t, t \pm 1, \cdots, t \pm H}
\end{equation}
where $J$ denotes the filter window with length $2H+1$, and the filter weights are given by two Gaussian functions as  
\begin{equation}\label{eq:denoising_weight} 
w_j^t = \frac{1}{z} e^{-\frac{|j-t|^2}{2\delta_d^2}} e^{-\frac{|y_j-y_t|^2}{2\delta_i^2}}, 
\end{equation}
where $\frac{1}{z}$ is a normalization factor, $\delta_d^2$ and $\delta_i^2$ are two parameters which control how smooth the output time series will be. 

After denoising, the decomposition model in Eq.~(\ref{eq:model}) is updated as 
\begin{align}\label{eq:updated_model} 
y_t' &= \tau_t + s_t + r_t'  \\
r_t' &= a_t + (n_t - \hat{n}_t)
\end{align}
where the $\hat{n}_t = y_t - y_t'$ is the filtered noise. 

\subsection{Trend Extraction}\label{subsection:detrend}


The joint learning of $\tau_t$ and $s_t$ in Eq.~(\ref{eq:updated_model}) is challenging. As the seasonality component is assumed to change slowly, we first perform seasonal difference operation for the denoised signal $y_t'$ to mitigate the seasonal effects, i.e.,
\begin{align} \notag
g_t &= \nabla_T y_t' = y_t' - y_{t-T}'  \notag \\
&=  \nabla_T \tau_t + \nabla_T s_t + \nabla_T r_t' \notag\\ 
&= \sum_{i=0}^{T-1} \nabla \tau_{t-i} + ( \nabla_T s_t + \nabla_T r_t'), \label{eq:season_diff}
\end{align}
where $\nabla_T x_t = x_t - x_{t-T}$ is the seasonal difference operation, and $\nabla x_t = x_t - x_{t-1}$ is the first order difference operation. 

Note that in Eq.~(\ref{eq:season_diff}), $\sum_{i=0}^{T-1} \nabla \tau_{t-i}$ dominates $g_t$ as we assume the seasonality difference operator on $s_t$ and $r_t'$ leads to significantly smaller values. Thus, we propose to recover the first order difference of trend signal $\nabla \tau_{t}$ from $g_t$ by minimizing the following weighted sum objective function
\begin{equation}\label{eq:trend_obj} 
\sum_{t=T+1}^{N} | g_t - \sum_{i=0}^{T-1} \nabla \tau_{t-i} | 
+ \lambda_1 \sum_{t=2}^{N} | \nabla \tau_{t} |  + \lambda_2 \sum_{t=3}^{N} | \nabla^2 \tau_{t} |,
\end{equation}
where $\nabla^2 x_t = \nabla (\nabla x_t) = x_t - 2x_{t-1} + x_{t-2}$ denotes the second order difference operation. The first term in Eq.~(\ref{eq:trend_obj}) corresponds to the empirical error using the LAD~\cite{lad:lasso} instead of the commonly used sum-of-squares loss function due to the its well-known robustness to outliers. Note that here we assume that $g_t \approx \sum_{i=0}^{T-1} \nabla \tau_{t-i}$. This assumption may not be true in the beginning for some $t$, but since the proposed framework employs an alternating algorithm to update trend $\tau_t$ and seasonality $s_t$ iteratively, later we can remove $\nabla_T s_t + \nabla_T r_t'$ from $g_t$ to make it hold. The second and third terms are first-order and second-order difference operator constraints for the trend component $\tau_t$, respectively. The second term assumes that the trend difference $\nabla \tau_{t}$ usually changes slowly but can also exhibit some abrupt level shifts; the third term assumes that the trends are smooth and piecewise linear such that  $\nabla^2 x_t = \nabla (\nabla x_t) = x_t - 2x_{t-1} + x_{t-2}$ are sparse~\cite{kim2009ell_1}. Thus, we expect the trend component $\tau_t$ can capture both abrupt level shift and gradual change.

The objective function \eqref{eq:trend_obj} can be rewritten in an equivalent matrix form as
\begin{equation}\label{eq:trend_obj_matrix} 
 || \mathbf{g} - \mathbf{M} \nabla \boldsymbol{\tau} ||_1
+ \lambda_1 || \nabla \boldsymbol{\tau} ||_1  + \lambda_2 ||  \mathbf{D} \nabla \boldsymbol{\tau} ||_1,
\end{equation}
where $||\mathbf{x}||_1 = \sum_i |x_i|$ denotes the $\ell_1$-norm of the vector $\mathbf{x}$, 
$\mathbf{g}$ and $\nabla \boldsymbol{\tau}$ are the corresponding vector forms as
\begin{equation}\label{eq:mathbf_g} 
\mathbf{g} = [g_{T+1}, g_{T+2}, \cdots, g_{N}]^T,
\end{equation}
\begin{equation}\label{eq:mathbf_tau} 
\nabla \boldsymbol{\tau} = [\nabla \tau_{2}, \nabla \tau_{3}, \cdots, \nabla \tau_{N}]^T,
\end{equation}
$\mathbf{M}$ and $\mathbf{D}$ are ${(N - T) \times (N - 1)}$ and ${(N - 2) \times (N - 1)}$ Toeplitz matrix, respectively, 
with the following forms
\vspace{4mm}
\newcommand\bovermat[2]{%
\makebox[0pt][l]{$\smash{\overbrace{\phantom{%
\begin{matrix}#2\end{matrix}}}^{\text{#1}}}$}#2}
\begin{align}\label{eq:matrix_M}  \notag
 \mathbf{M} \!= \!\!
 \begin{bmatrix}
 \bovermat{\textit{T} ones}{
 1 &  \cdots & 1}\\ 
 &1 &  \cdots & 1\\
 &&\ddots\\
 &&1 & \cdots & 1
 \end{bmatrix}\!\!,~
 \mathbf{D} \!= \!\! 
 \begin{bmatrix}
 1 & -1\\
 &1&-1\\
 &&\ddots\\
 &&1&-1
 \end{bmatrix}.
\end{align}

To facilitate the process of solving the above optimization problem, we further formulate the three $\ell_1$-norms in Eq.~(\ref{eq:trend_obj_matrix}) as a single $\ell_1$-norm, i.e.,
 \begin{equation}\label{eq:trend_obj_one_L1} 
 || \mathbf{P} \nabla \boldsymbol{\tau} -  \mathbf{q}  ||_1,
 \end{equation}
where the matrix $\mathbf{P}$  and vector $\mathbf{q}$ are 
\begin{equation}\label{eq:matrix_P_q}  \notag
\mathbf{P}= 
\begin{bmatrix}
\mathbf{M}_{(N - T) \times (N - 1)} \\
\lambda_1 \mathbf{I}_{(N - 1) \times (N - 1)}\\
\lambda_2 \mathbf{D}_{(N - 2) \times (N - 1)} \\
\end{bmatrix}, 
\mathbf{q}= 
\begin{bmatrix}
\mathbf{g}_{(N - T) \times 1}\\
\mathbf{0}_{(2N - 3) \times 1}\\
\end{bmatrix}.
\end{equation} 
The minimization of the single $\ell_1$-norm in Eq.~(\ref{eq:trend_obj_one_L1}) is equivalent to the linear program as follows
\begin{equation}\label{eq:trend_LP_opt} 
\begin{matrix}
\displaystyle 
\min & 
\begin{bmatrix}
\mathbf{0}\\
\mathbf{1} \\
\end{bmatrix}^T
\begin{bmatrix}
\nabla \boldsymbol{\tau}\\
\mathbf{v} \\
\end{bmatrix}\vspace{2mm}\\
\textrm{s.t.} & 
\begin{bmatrix}
\mathbf{P} & -\mathbf{I}\\
-\mathbf{P} & -\mathbf{I}\\
\end{bmatrix}
\begin{bmatrix}
\nabla \boldsymbol{\tau}\\
\mathbf{v} \\
\end{bmatrix}
\leq
\begin{bmatrix}
\mathbf{q}\\
-\mathbf{q} \\
\end{bmatrix},
\end{matrix}
\end{equation}
where $\mathbf{v}$ is auxiliary vector variable. Let denote the output of the above optimization is $[\nabla \tilde{\boldsymbol{\tau}}, \tilde{\mathbf{v}}]^T$ with $\nabla \tilde{\boldsymbol{\tau}} = [\nabla \tilde{\tau}_{2}, \nabla \tilde{\tau}_{3}, \cdots, \nabla \tilde{\tau}_{N}]^T$, and further assume $\tilde{\tau}_{1} =  {\tau}_{1}$ (which will be be estimated later). Then, we can get the relative trend output  based on $ {\tau}_{1}$  as
\begin{equation}\label{eq:relative_trend} 
\tilde{\tau}_{t}^r = \tilde{\tau}_{t} - {\tau}_{1} = \tilde{\tau}_{t} - \tilde{\tau}_{1}  = 
\left\{\begin{array}{lcl}
0,                                             		   & t = 1\\
\sum_{i=2}^{t} \nabla \tilde{\tau}_{i},  & t \geq 2  
\end{array}\right.{\!\!\!\!}.
\end{equation}

Once obtaining the relative trend from the denoised time series, the decomposition model is updated as 
\begin{equation}\label{eq:signal_remove_trend} 
y_t'' = y_t' - \tilde{\tau}_{t}^r = s_t + {\tau}_{1} + r_t'',
\end{equation}
\begin{equation}\label{eq:rt_3prime} 
r_t'' = a_t + (n_t - \hat{n}_t) + (\tau_t - \tilde{\tau}_{t}).
\end{equation}

\subsection{Seasonality Extraction}\label{subsection:deseason} 
After removing the relative trend component, $y_i''$ can be considered as a ``contaminated seasonality''. In addition, seasonality shift makes the estimation of $s_i$ even more difficult. Traditional seasonality extraction methods only considers a subsequence $y_{t-KT}, y_{t-(K-1)T}, \cdots, y_{t-T}$ (or associated sequences) to estimate $s_t$, where $T$ is the period length. However, this approach fails when seasonality shift happens. 

In order to accurately estimate the seasonality component, we propose a non-local seasonal filtering. In the non-local seasonal filtering, instead of only considering $y_{t-KT}, \cdots, y_{t-T}$, we consider $K$ neighborhoods centered at $y_{t-KT}, \cdots, y_{t-T}$, respectively. Specifically, for $y_{t-kT}''$, its neighborhood consists of $2H+1$ neighbors $y_{t-kT-H}'', y_{t-kT-H+1}'', \cdots, y_{t-kT}'', y_{t-kT+1}'', \cdots, y_{t-kT+H}''$. Furthermore, we model the seasonality component $s_t$ as a weighted linear combination of $y_j''$ where $y_j''$ is in the neighborhood defined above. The weight between $y_t''$ and $y_j''$ depends not only in how far they are in the time dimension (i.e., the difference of their indices $t$ and $j$), but also depends on how close $y_t''$ and $y_j''$ are. Intuitively, the points in the neighborhood with similar seasonality to $y_t$ will be given a larger weight. In this way, we automatically find the points with most similar seasonality and solve the seasonality shift problem. In addition, abnormal points will be given smaller weights in our definition and makes the non-local seasonal filtering robust to outliers.


Mathematically, the operation of seasonality extraction by non-local seasonal filtering is formulated as
\begin{equation}\label{eq:relative_season} 
\tilde{s}_t  =\sum_{(t',j) \in \Omega} w_{(t',j)}^t y_j''
\end{equation}
where the $w_{(t',j)}^t$ and $\Omega$ are defined as
\begin{align}\label{eq:weight_irregular_from_season} 
w_{(t',j)}^t &= \frac{1}{z} e^{-\frac{|j-t'|^2}{2\delta_d^2}} e^{-\frac{|y_j''-y_{t'}''|^2}{2\delta_i^2}} \\
\Omega    &= \{(t',j) | (t'=t-k \times T, j = t' \pm h)\} \\ \notag
                &~~~~~k=1,2,\cdots,K; ~~h=0,1,\cdots, H
\end{align}
by considering previous $K$ seasonal neighborhoods where each neighborhood contains $2H+1$ points. 

\begin{figure}[!b]
    \centering
    \subfigure[Outlier robustness]{
        \includegraphics[width=0.41\linewidth]{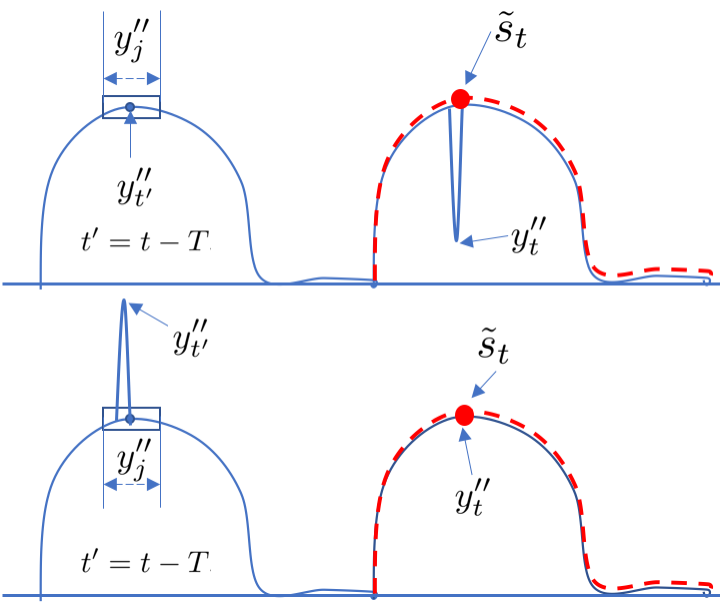}
        }
    \subfigure[Season shift adaptation]{
        \includegraphics[width=0.52\linewidth]{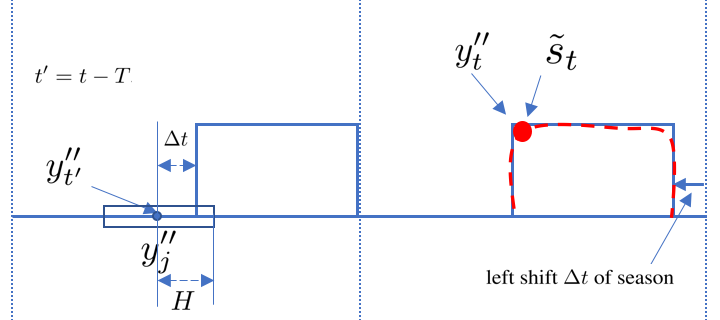}
        }
    \caption{Robust and adaptive properties of the non-local seasonal filtering (red curve denotes the extracted seasonal signal).}
    \label{fig:3_season_spikeshift}
\end{figure}

To illustrate the robustness of our non-local seasonal filtering to outliers, we give an example in Figure~\ref{fig:3_season_spikeshift}(a). Whether there is outlier at current time point $t$ (dip), or at historical time point around $t-T$ (spike), the filtered output $\tilde{s}_t$ (red curve) would not be affected as show in Figure \ref{fig:3_season_spikeshift}(a). Figure \ref{fig:3_season_spikeshift}(b) illustrates why seasonality shift is overcome by the non-local seasonal filtering. As shown in the red curve in Figure \ref{fig:3_season_spikeshift}(b), when there is $\Delta t$ shift in the season pattern, as long as the previous seasonal neighborhoods used in the non-local seasonal filter satisfy $H > \Delta t$, the extracted season can follow this shift.

After removing the season signal, the remainder signal is 
\begin{align}\label{eq:irregular_from_season} \notag
&r_t''' = y_t'' - \tilde{s}_t = a_t + (n_t - \hat{n}_t) + (\tau_t - \tilde{\tau}_{t})   + (s_t - \tilde{s}_t).
\end{align}

\subsection{Final Adjustment}\label{subsection:adjust} 

In order to make the seasonal-trend decomposition unique, we need to ensure that all seasonality components in a period sums to zero, i.e., $\sum_{i=j}^{i=j+T-1} s_i = 0$. To this end, we adjust the obtained seasonality component from Eq.~(\ref{eq:relative_season}) by removing its mean value, which also corresponds to the estimation of the trend point ${\tau}_1$. Formally, we have 
\begin{equation}\label{eq:est_tau_1} 
\hat{\tau}_1 = \frac{1}{T \lfloor N/T \rfloor} \sum_{t=1}^{{T \lfloor N/T \rfloor} } \tilde{s}_t.
\end{equation}
Therefore, the estimates of trend and seasonal components are updated as follows:
\begin{equation}\label{eq:final_season} 
\hat{s}_t = \tilde{s}_t - \hat{\tau}_1,
\end{equation}
\begin{equation}\label{eq:final_trend} 
\hat{\tau}_t = \tilde{\tau}_t^r + \hat{\tau}_1.
\end{equation}
And the estimate of remainder can be obtained as
\begin{equation}\label{eq:final_irregular} 
\hat{r}_t = r_t''' + \hat{n}_t \quad  \text{or}  \quad \hat{r}_t = y_t - \hat{s}_t - \hat{\tau}_t.
\end{equation}
Furthermore, we can derive different components of the estimated $\hat{r}_t$, i.e.,
\begin{equation}\label{eq:final_irregular_components} 
\hat{r}_t = 
\left\{\begin{array}{lcl}
 a_t + {n}_t + (s_t - \hat{s}_t) +  ({\tau}_{1} - \hat{\tau}_{1}) ,    & t = 1 \\
 a_t + {n}_t + (s_t - \hat{s}_t) +  (\tau_t - \tilde{\tau}_{t}) ,    & t \geq 2  
\end{array}\right.
\end{equation}

 
Eq.~\eqref{eq:final_irregular_components} indicates that the remainder signal $\hat{r}_t$ may contain residual seasonal component $(s_t - \hat{s}_t)$ and trend component $(\tau_t - \tilde{\tau}_{t})$. Also in the trend estimation step, we assume $\sum_{i=0}^{T-1} \nabla \tau_{t-i}$ can be approximated using $g_t$. Similar to alternating algorithm, after we obtained better estimates of $\tau_t$, $s_t$, and $r_t$, we can repeat the above four steps to get more accurate estimates of the trend, seasonality, and remainder components. Formally, the procedure of the RobustSTL algorithm is summarized in Algorithm \ref{alg:RobustSTL}.

\floatname{algorithm}{Algorithm}
\renewcommand{\algorithmicrequire}{\textbf{Input:}}
\renewcommand{\algorithmicensure}{\textbf{Output:}}
\newcommand{\INDSTATE}[1][1]{\STATE\hspace{#1\algorithmicindent}}

\begin{algorithm}
\caption{RobustSTL Algorithm Summary}
\label{alg:RobustSTL}
\begin{algorithmic}
\REQUIRE $y_t$, parameter configurations.
\ENSURE $\hat{\tau}_t, \hat{s}_t, \hat{r}_t$

\STATE \textbf{Step 1:} \text{Denoise input signal using bilateral filter}
\INDSTATE[1] $w_j^t = \frac{1}{z} e^{-\frac{|j-t|^2}{2\delta_d^2}} e^{-\frac{|y_j-y_t|^2}{2\delta_i^2}}, ~~ y_t' = \sum_{j \in J} w_j^t y_j$

\STATE \textbf{Step 2:} \text{Obtain relative trend from $\ell_1$ sparse model}
\INDSTATE[1] $ \nabla \tilde{\boldsymbol{\tau}}  = \arg\min_{\nabla \boldsymbol{\tau}} \!|| \mathbf{P} \nabla \boldsymbol{\tau} -  \mathbf{q}  ||_1 \text{(see Eq. \eqref{eq:trend_obj}, \eqref{eq:trend_obj_matrix}, \eqref{eq:trend_obj_one_L1})} $ 
\INDSTATE[1] $\tilde{\tau}_{t}^r = 
\left\{\begin{array}{lcl}
0,                                             		   & t = 1\\
\sum_{i=2}^{t} \nabla \tilde{\tau}_{i},  & t \geq 2  
\end{array}\right.{\!\!\!\!},$ 
\INDSTATE[1] $y_t'' = y_t' - \tilde{\tau}_{t}^r$

\STATE \textbf{Step 3:} \text{Obtain season using non-local seasonal filtering}
\INDSTATE[1] $w_{(t',j)}^t = \frac{1}{z} e^{-\frac{|j-t'|^2}{2\delta_d^2}} e^{-\frac{|y_j''-y_{t'}''|^2}{2\delta_i^2}}$
\INDSTATE[1] $\tilde{s}_t  =\sum_{(t',j) \in \Omega} w_{(t',j)}^t y_j''$

\STATE \textbf{Step 4:} \text{Adjust trend and season}
\INDSTATE[1] $\hat{\tau}_1 = \frac{1}{T \lfloor N/T \rfloor} \sum_{t=1}^{{T \lfloor N/T \rfloor} } \tilde{s}_t$
\INDSTATE[1] $\hat{\tau}_t = \tilde{\tau}_t^r + \hat{\tau}_1, ~ \hat{s}_t = \tilde{s}_t - \hat{\tau}_1, ~ \hat{r}_t = y_t - \hat{s}_t - \hat{\tau}_t$

\STATE \textbf{Step 5:} \text{Repeat Steps 1-4 for $\hat{r}_t$ until convergence}
\end{algorithmic}
\end{algorithm}

\section{Experiments}
We conduct experiments to demonstrate the effectiveness of the proposed RobustSTL algorithm on both synthetic and real-world datasets.

\subsection{Baseline Algorithms}
We use the following three state-of-the-art baseline algorithms for comparison purpose:
\begin{itemize}
    \item \textit{Standard STL}: It decomposes the signal into seasonality, trend, and remainder based on Loess in an iterative manner. 
    \item \textit{TBATS}: It decomposes the signal into trend, level, seasonality, and remainder. The trend and level are jointly together to represent the real trend. 
    \item \textit{STR}: It assumes the continuity in both trend and seasonality signal. 
\end{itemize}
To test the above three algorithms, we use R functions \texttt{stl, tbats, AutoSTR} from R packages \texttt{forecast\footnote{https://cran.r-project.org/web/packages/forecast/index.html}} and \texttt{stR\footnote{https://cran.r-project.org/web/packages/stR/index.html}}. We implement our own RobustSTL algorithm in Python, where the linear program (see Eqs. \eqref{eq:trend_obj_one_L1} and \eqref{eq:trend_LP_opt}) in trend extraction is solved using CVXOPT $\ell_1$-norm approximation\footnote{https://cvxopt.org/examples/mlbook/l1.html}.

\subsection{Experiments on Synthetic Data}
\subsubsection{Dataset}

\begin{figure}[!htb]
    \centering
    \includegraphics[width=1.0\linewidth]{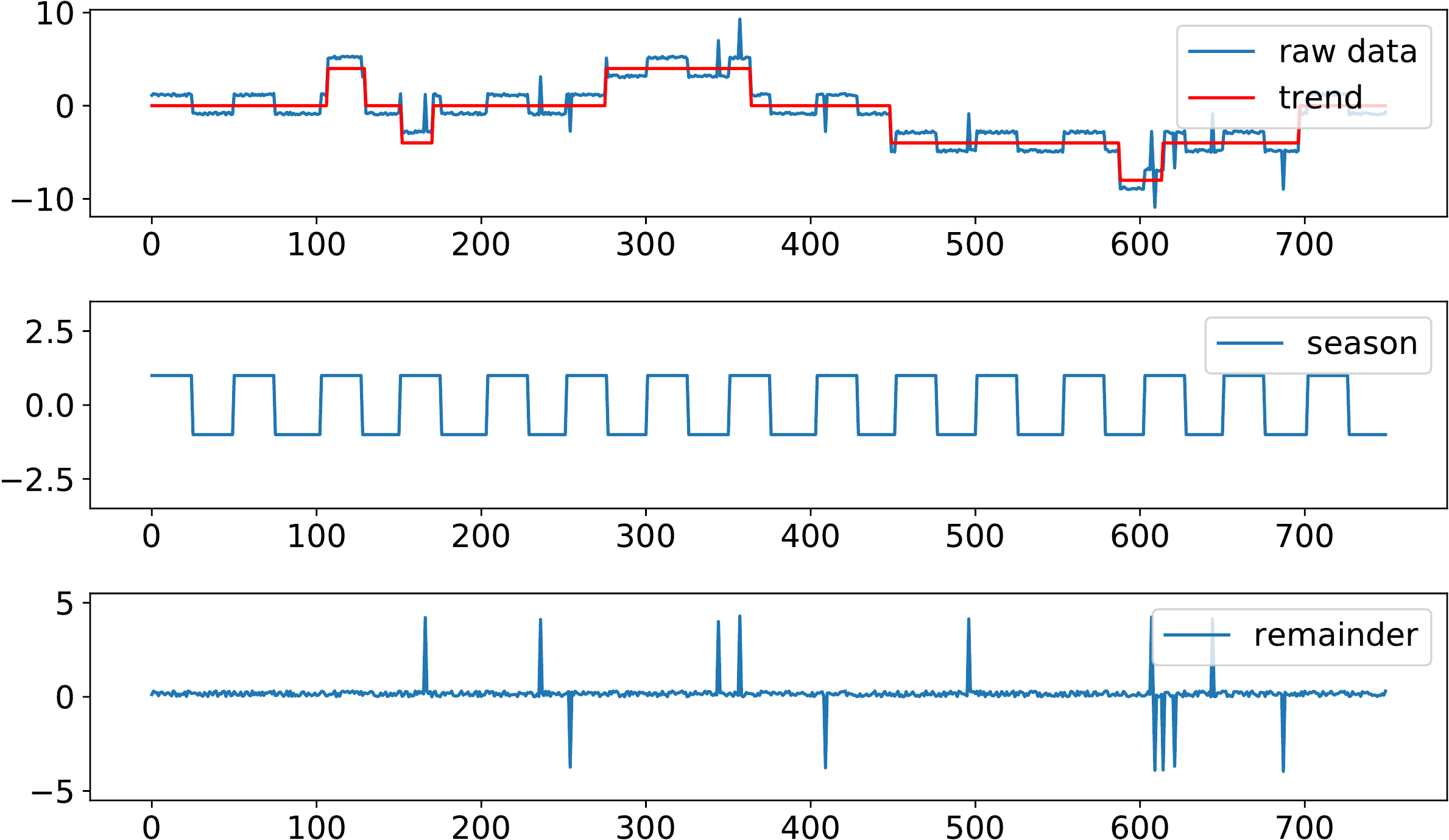}
    \caption{Generated synthetic data where the top subplot represents the raw data and the trend, the middle subplot represents the seasonality, and the bottom subplot represents the noise and anomalies.}
    \label{fig:4-origin-synthetic}
\end{figure}

To generate the synthetic dataset, we incorporate complex season/trend shifts, anomalies, and noise to simulate real-world scenarios as shown in Figure~\ref{fig:4-origin-synthetic}. We first generate seasonal signal using a square wave with minor random seasonal shifts in the horizontal axis. The seasonal period is set to 50 and a total of 15 periods are generated. Then, we add trend signal with 10 random abrupt level changes and 14 spikes and dips as anomalies. The noise is added by zero mean Gaussian with $0.1$ variance.

\begin{figure*}[!h]
    \centering
    \subfigure[RobustSTL]{
        \includegraphics[width=.44\linewidth]{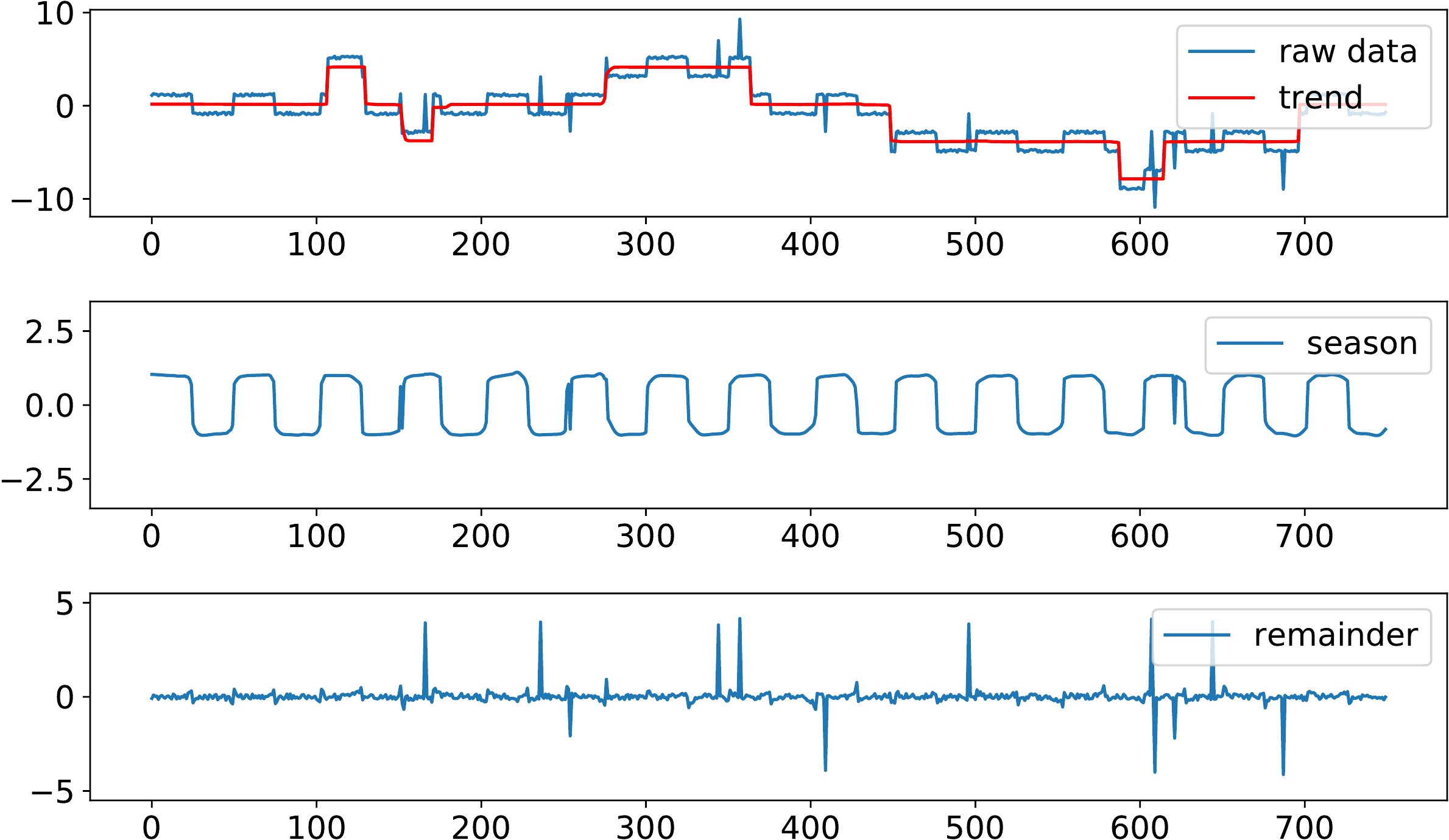}
        }
    \subfigure[Standard STL]{
        \includegraphics[width=.44\linewidth]{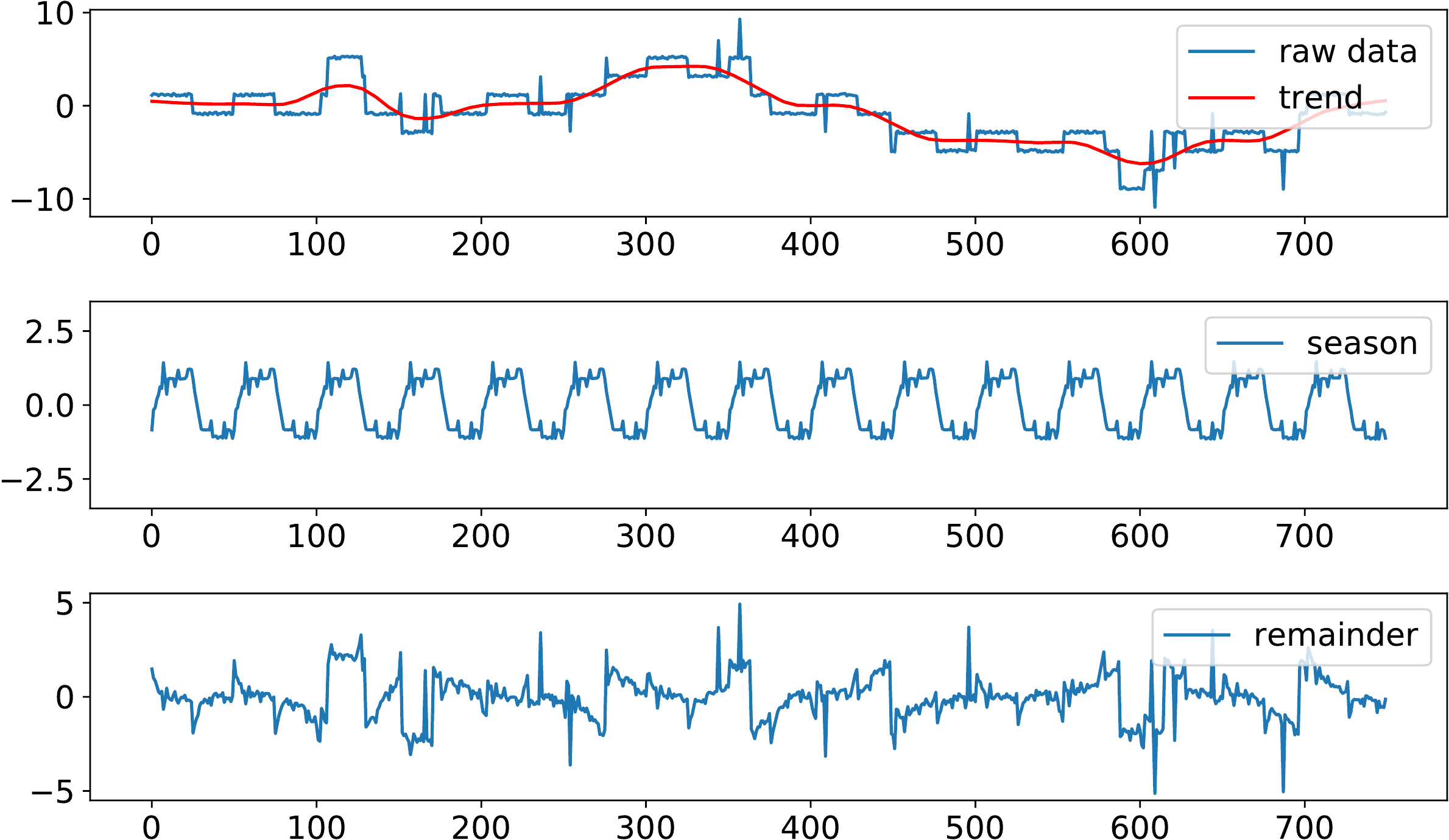}
        }
    \subfigure[TBATS]{
        \includegraphics[width=.44\linewidth]{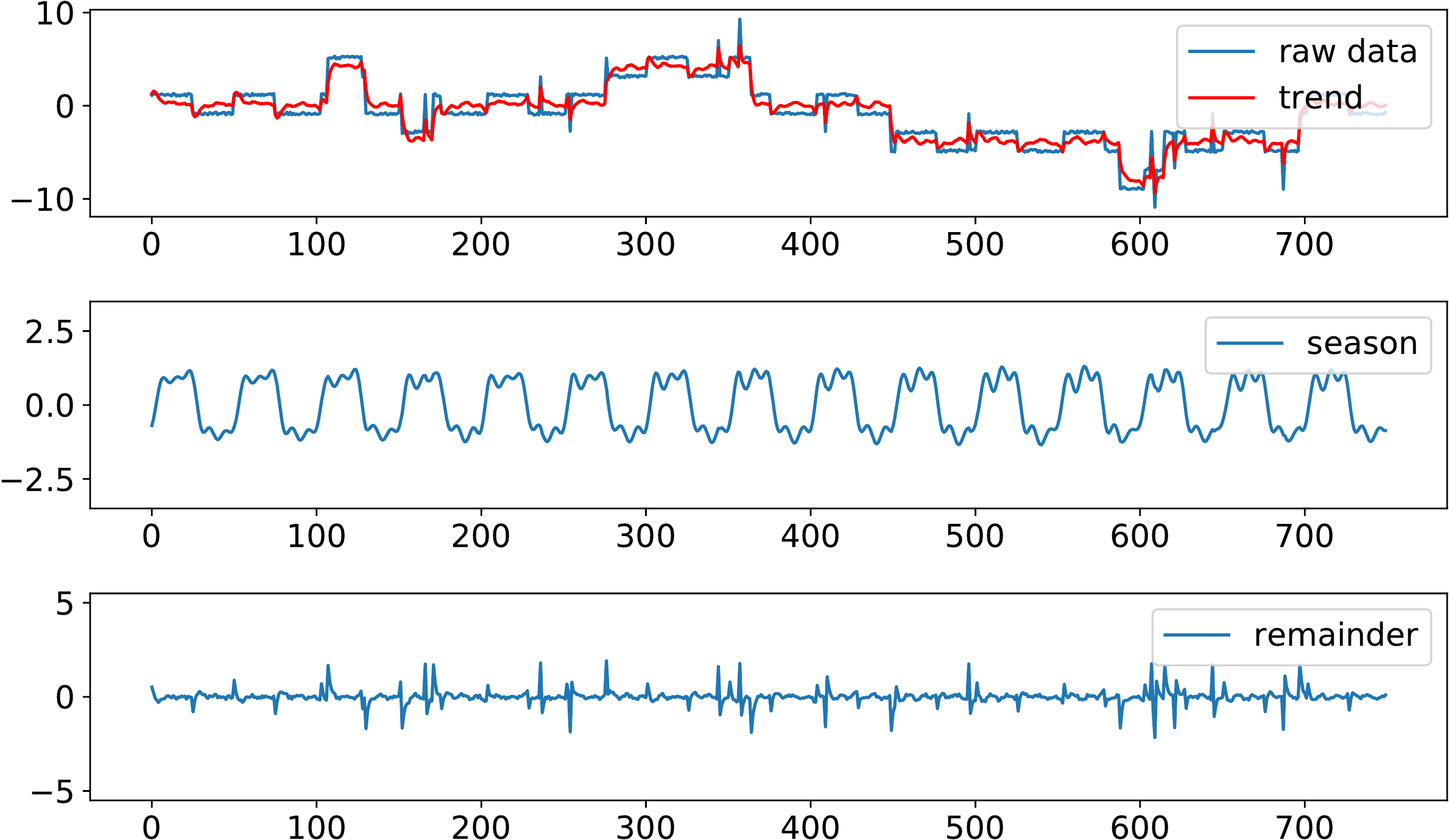}
        }
    \subfigure[STR]{
        \includegraphics[width=.44\linewidth]{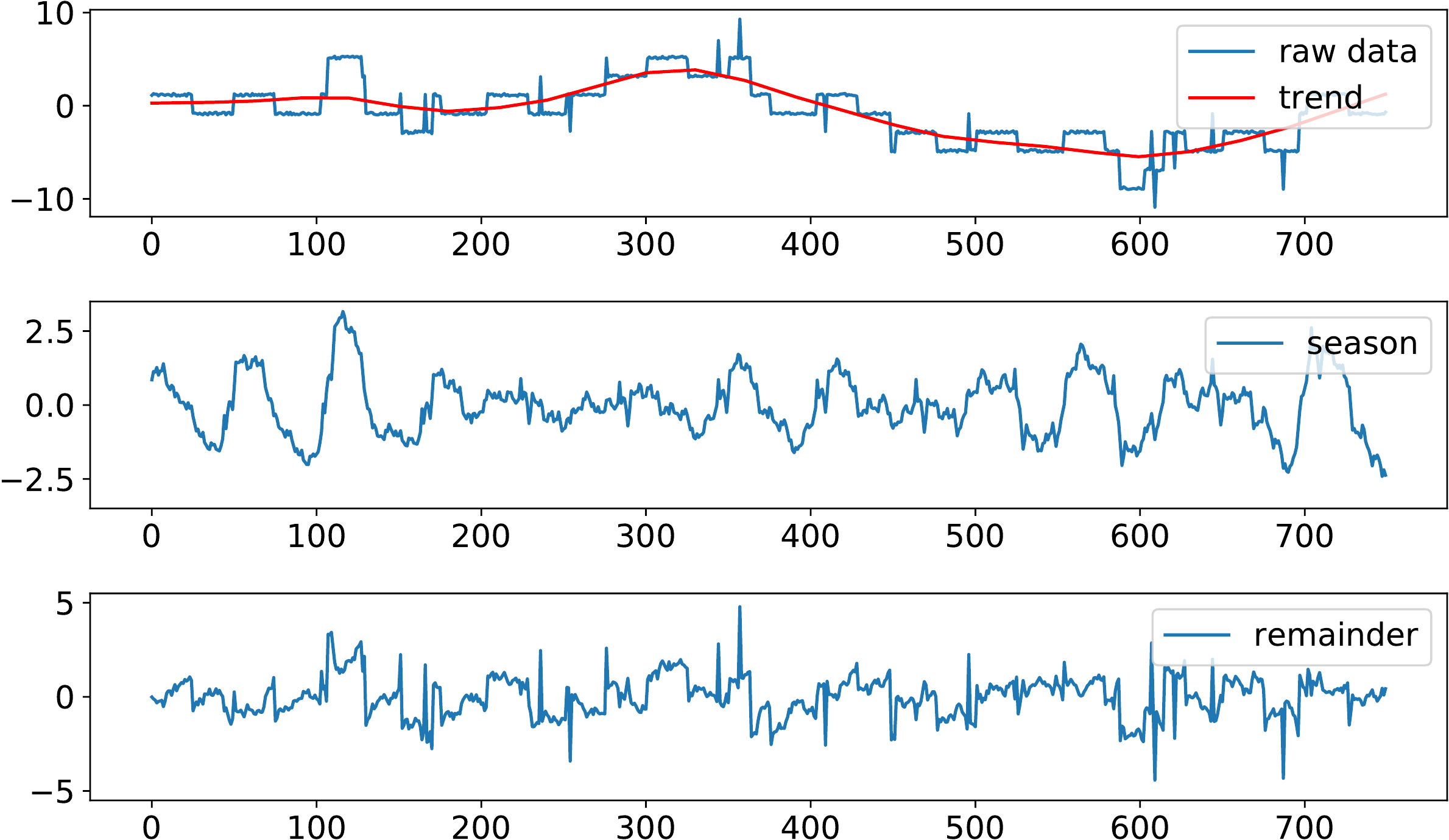}
        }
    \caption{Decomposition results on synthetic data for a) Robust STL; b) Standard STL; c) TBATS; and d) STR.}
    \label{fig:4-synthetic-results}
\end{figure*}
\subsubsection{Experimental Settings}

For the three baseline algorithms STL, TBATS, and STR, their parameters are optimized using cross-validation. For our proposed RobustSTL algorithm, we set the regularization coefficients $\lambda_1=10, \lambda_2=0.5$ to control the signal smoothness in the trend extraction, and set the neighborhood parameters $K=2, H=5$ in the seasonality extraction to handle the seasonality shift.

\subsubsection{Decomposition Results}

Figure~\ref{fig:4-synthetic-results} shows the decomposition results for four algorithms. The standard STL is affected by anomalies leading to rough patterns in the seasonality component. The trend component also tends to be smooth and loses the abrupt patterns. TBATS is able to respond to the abrupt changes and level shifts in trend, however, the trend is affected by the anomaly signals. It also obtains almost the same seasonality pattern for all periods. Meanwhile, STR does not assume strong repeated patterns for seasonality and is robust to outliers. However, it cannot handle the abrupt changes of trend. By contrast, the proposed RobustSTL algorithm is able to separate trend, seasonality, and remainders successfully, which are all very close to the original synthetic signals. 

\begin{table}[htb]
\caption{Comparison of MSE and MAE for trend and seasonality components of different decomposition algorithms.}\label{tab:trend_season_MSE_MAE}
\begin{tabular}{l|c|c|c|c}
\hline
\multirow{2}{*}{Algorithms}& \multicolumn{2}{c|}{Trend} & \multicolumn{2}{c}{Season} \\\cline{2-5}
              &   MSE       &    MAE    &    MSE    &   MAE   \\\hline\hline
STR           &     1.6761  &  0.9588   &   0.8474  & 0.7508  \\\hline
Standard STL &     0.9496  &  0.6806   &   0.1915  & 0.3004  \\\hline
TBATS         &     0.3150  &  0.3637   &   0.1839  & 0.2770  \\\hline
RobustSTL     &     \textbf{0.0530}  &  \textbf{0.1627}   &   \textbf{0.0265}  & \textbf{0.0750}  \\ \hline    
\end{tabular}
\end{table}

To evaluate the performance quantitatively, we also compare mean squared error (MSE) and mean absolute error (MAE) between the true trend/season in the synthetic dataset and the extracted trend/season from four decomposition algorithms, which is summarized in Table ~\ref{tab:trend_season_MSE_MAE}. It can be observed that our RobustSTL algorithm achieves much better results than STL, STR, and TBATS algorithms.

\begin{figure*}[!h]
    \centering
    \subfigure[RobustSTL]{
        \includegraphics[width=.45\linewidth]{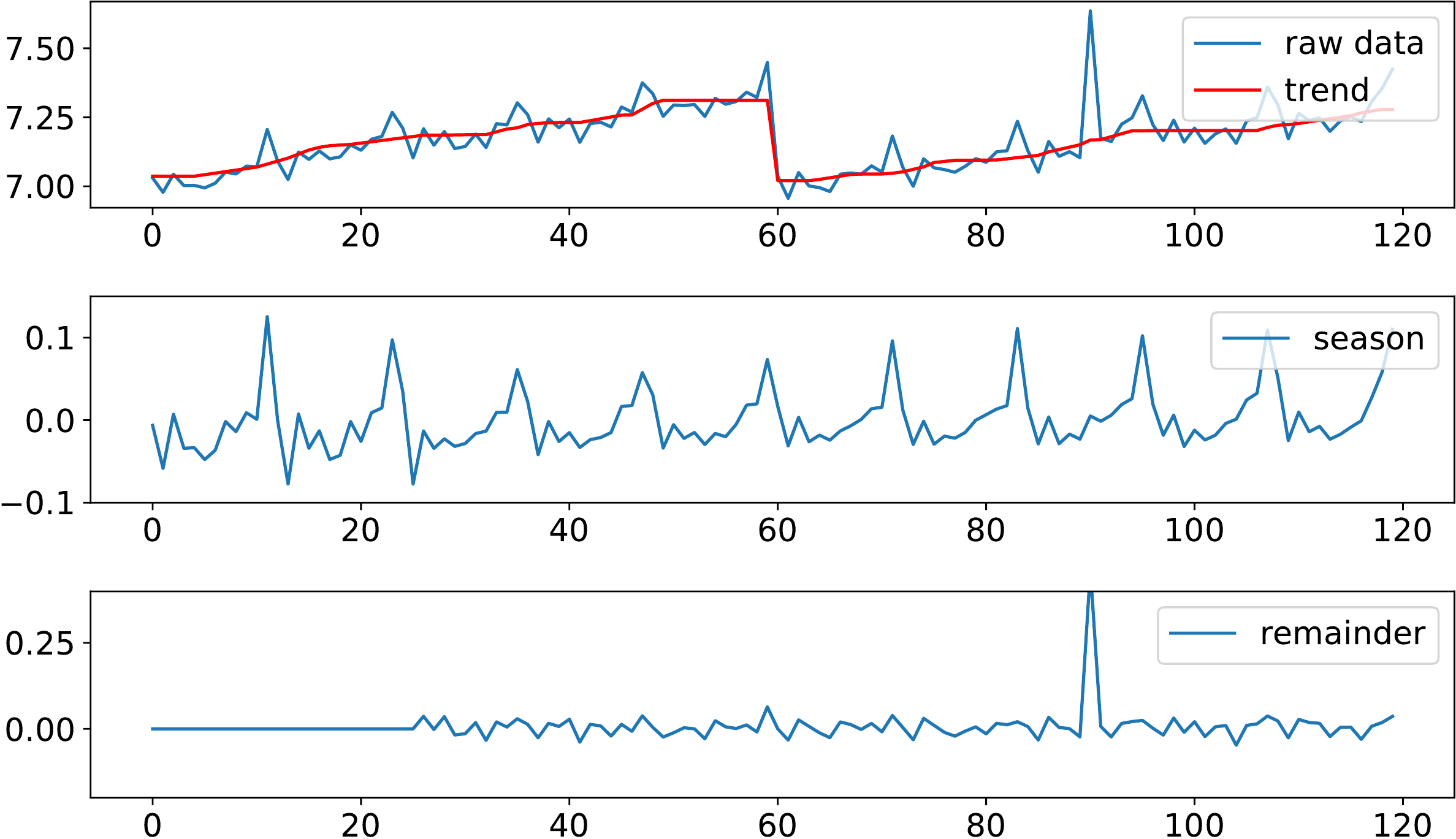}
        }
    \subfigure[Standard STL]{
        \includegraphics[width=.45\linewidth]{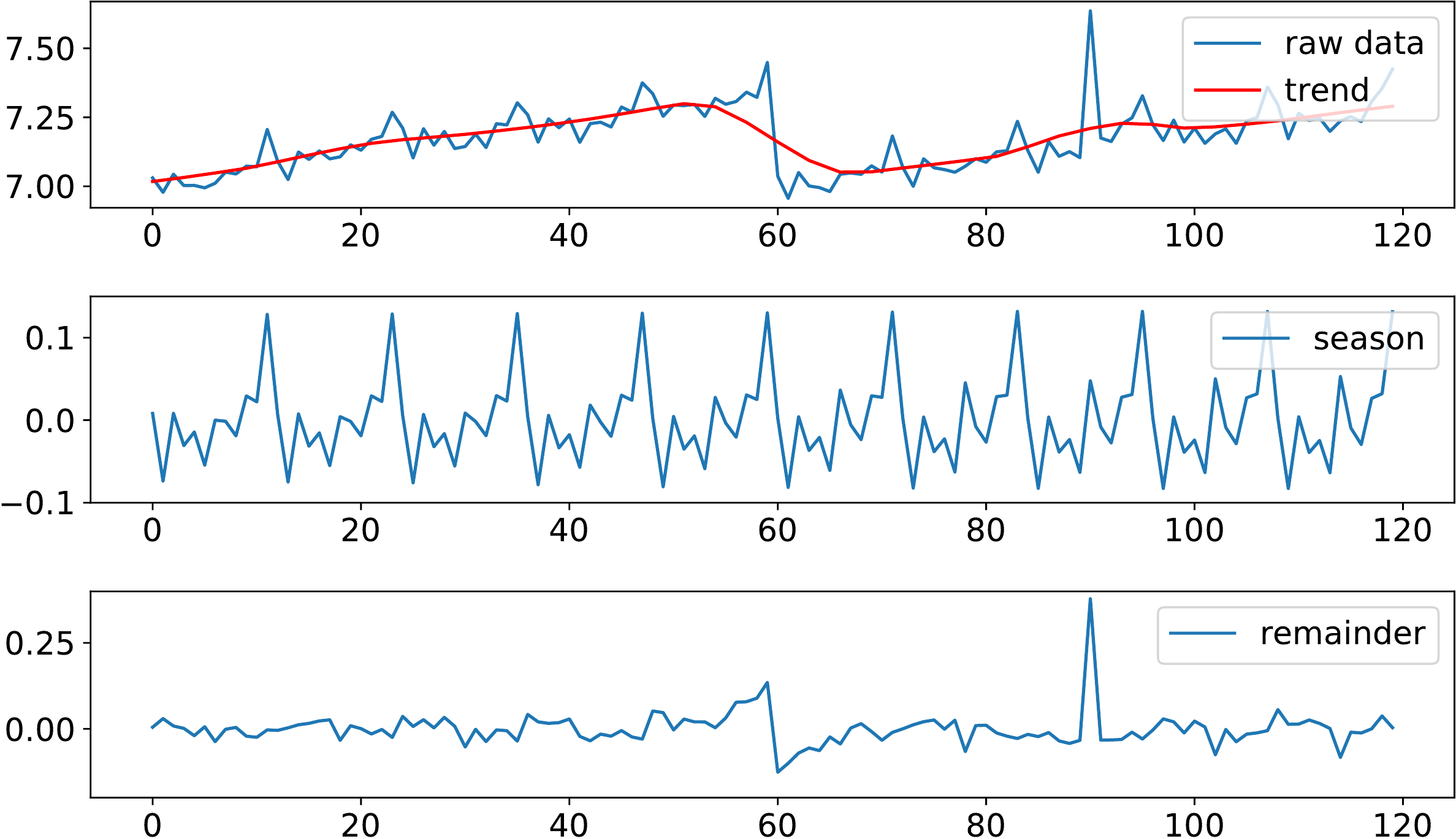}
        }
    \subfigure[TBATS]{
        \includegraphics[width=.45\linewidth]{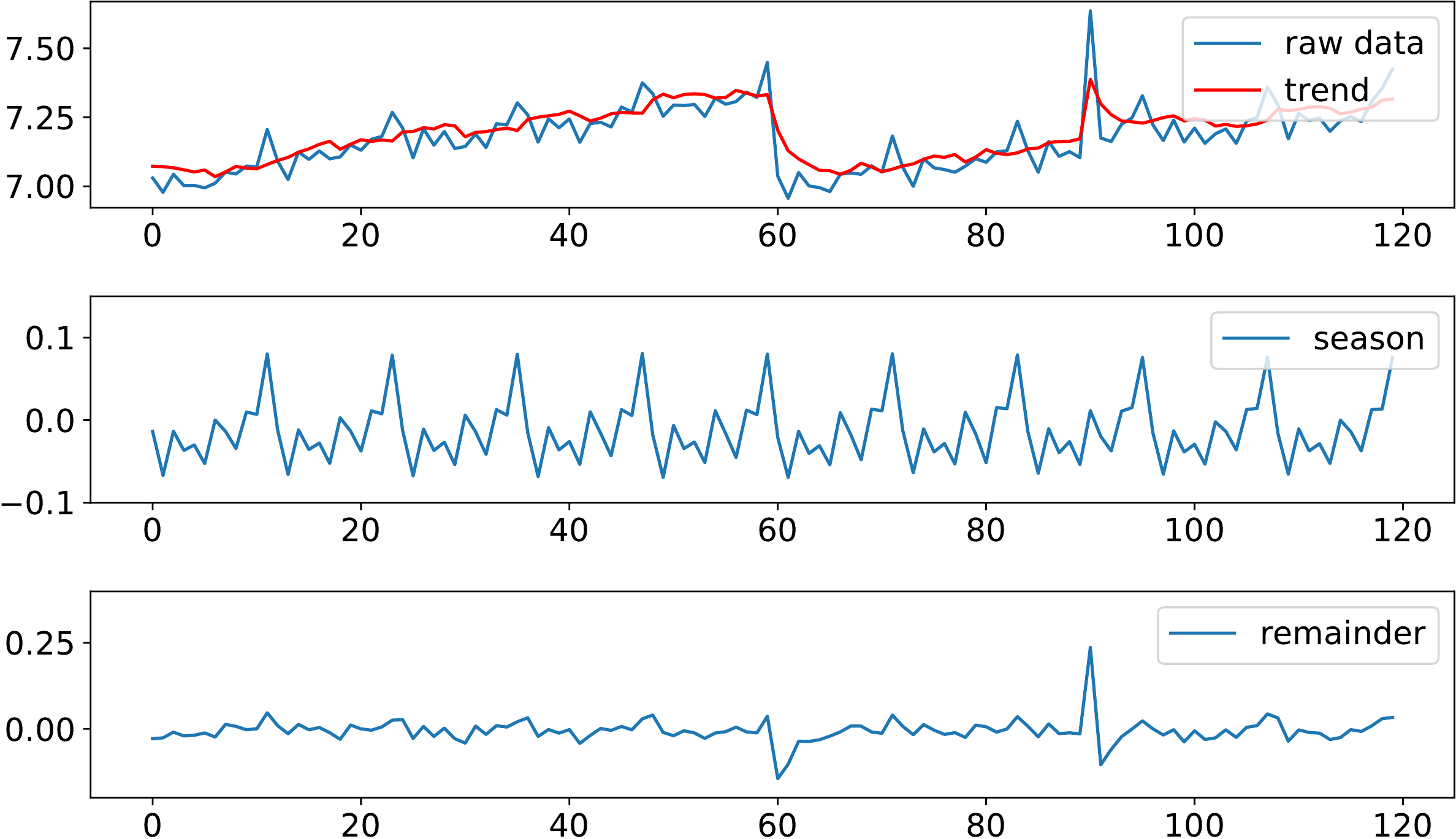}
        }
    \subfigure[STR]{
        \includegraphics[width=.45\linewidth]{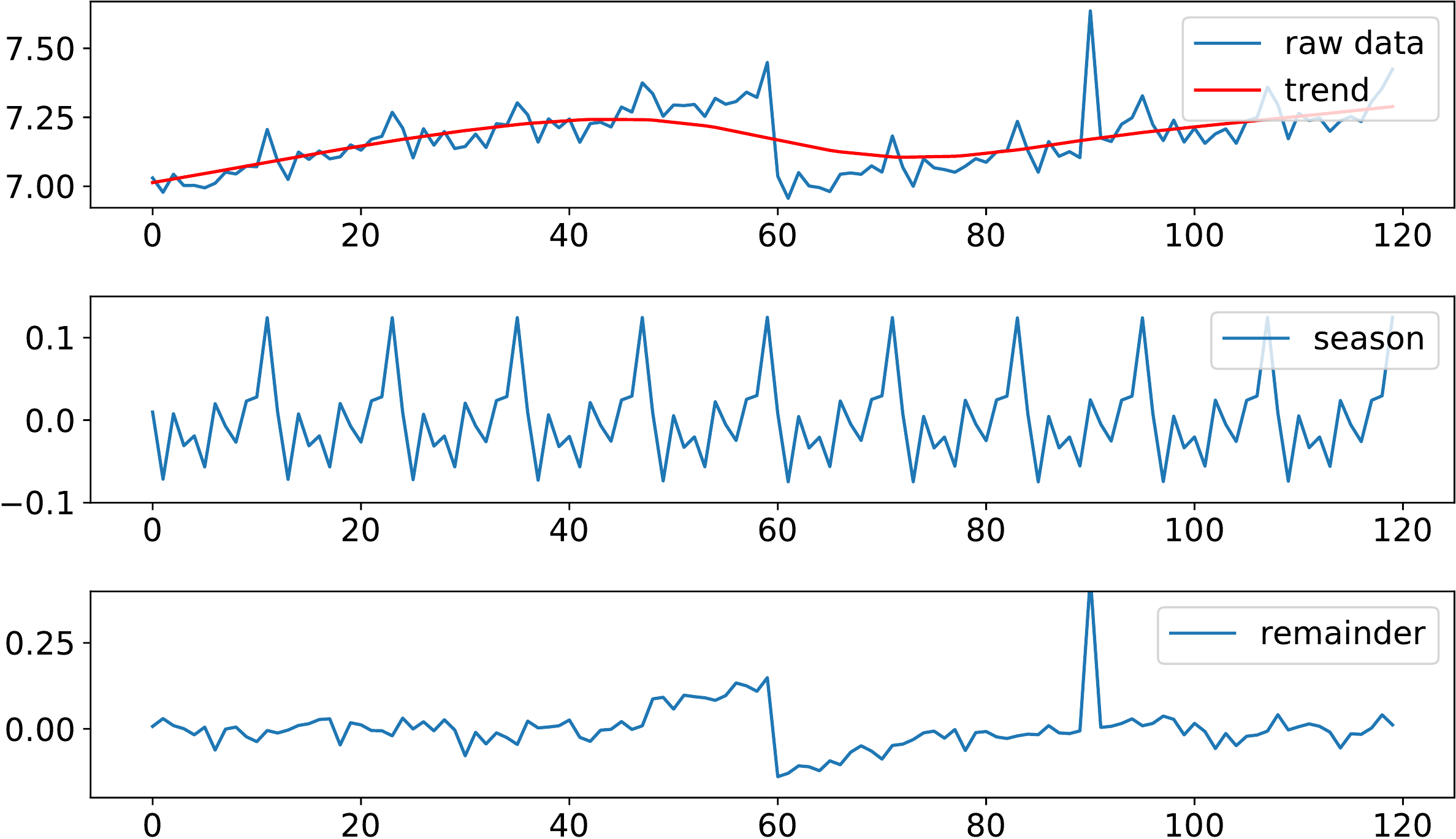}
        }
    \caption{Decomposition results on real dataset 1 for a) RobustSTL; b) Standard STL; c) TBATS and d) STR.}
    \label{fig:4-grocery-results}
\end{figure*}
\begin{figure*}[!h]
    \centering
    \subfigure[RobustSTL]{
        \includegraphics[width=.31\linewidth]{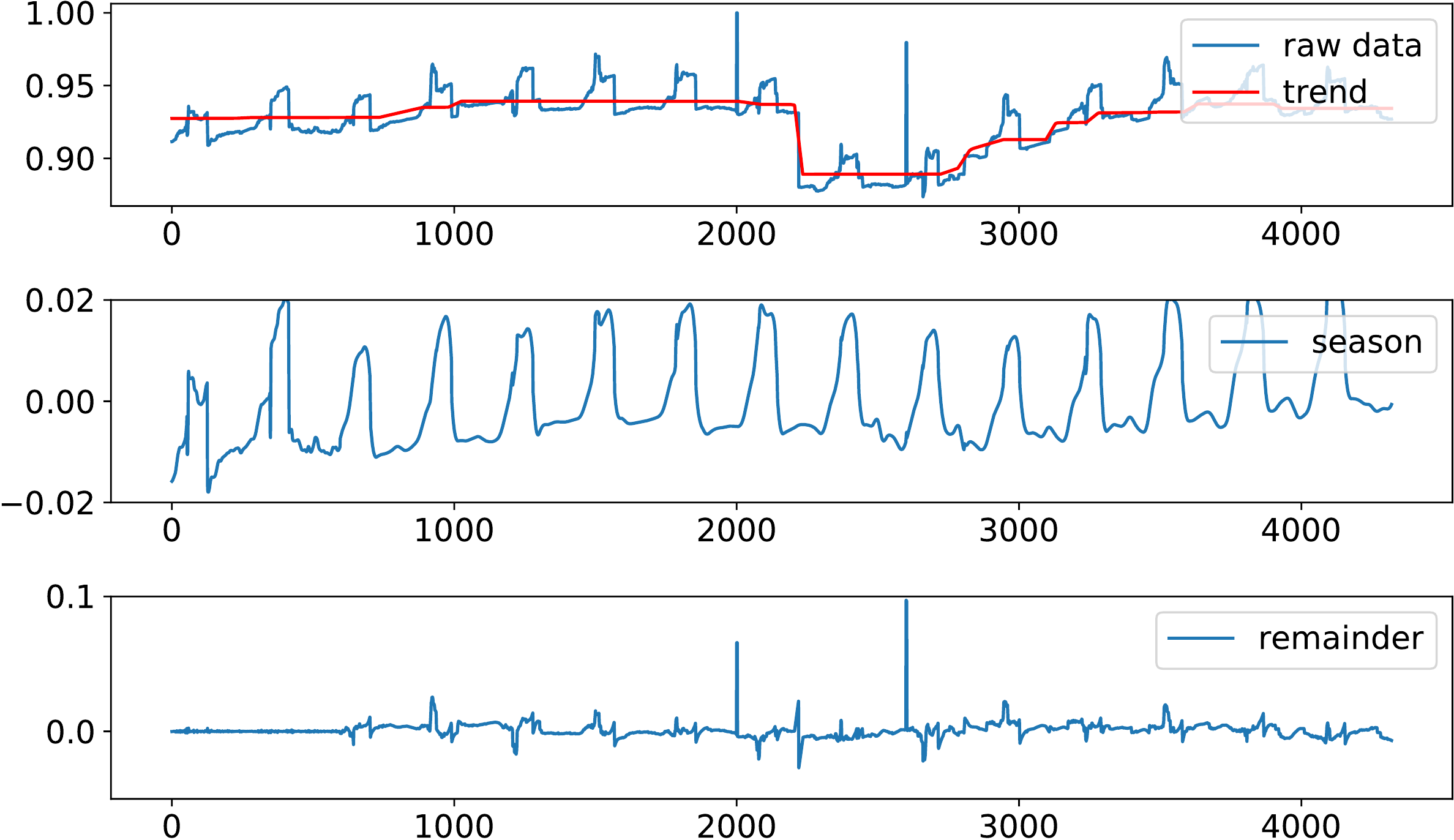}
        }
    \subfigure[Standard STL]{
        \includegraphics[width=.31\linewidth]{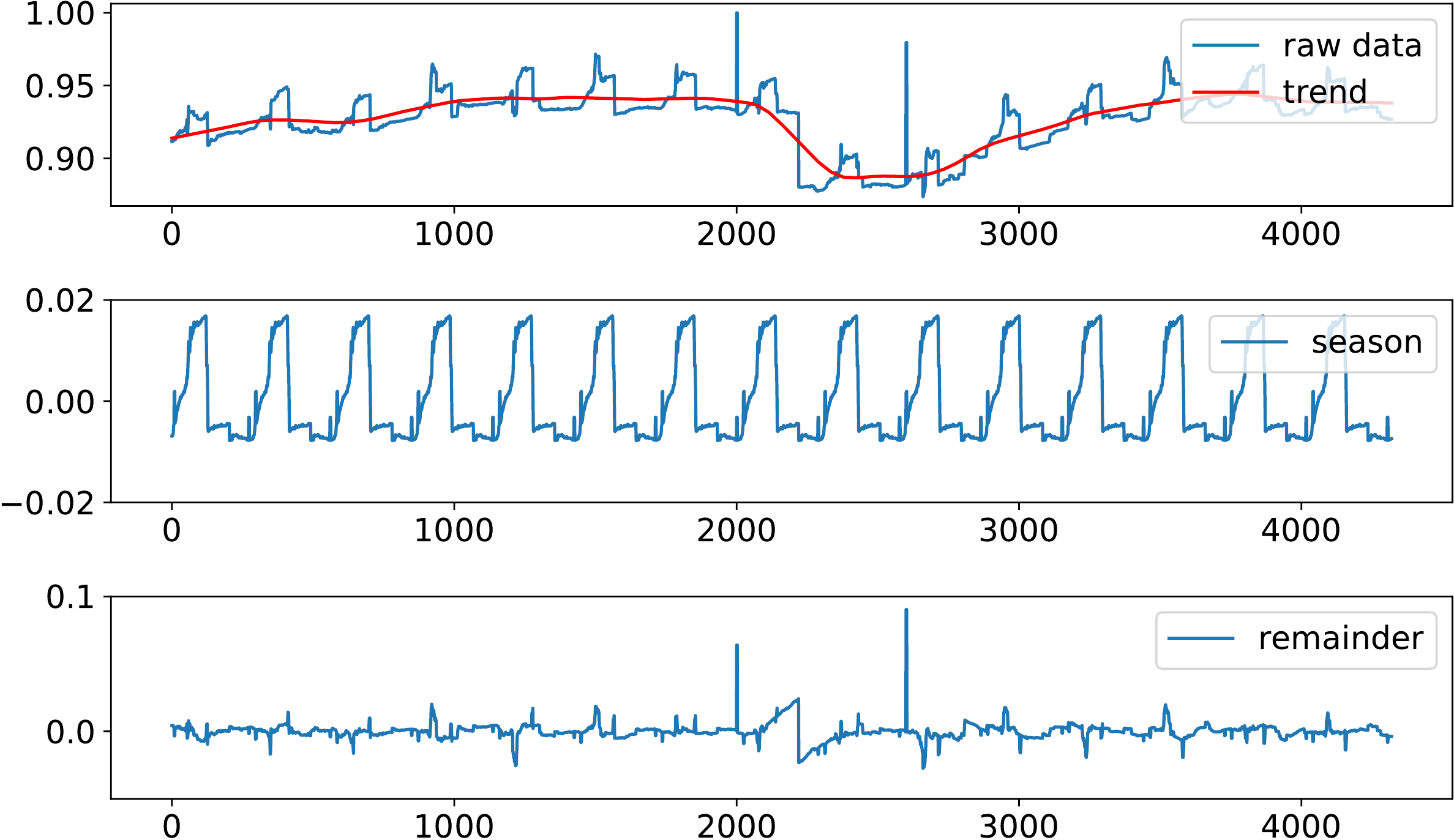}
        }
    \subfigure[TBATS]{
        \includegraphics[width=.31\linewidth]{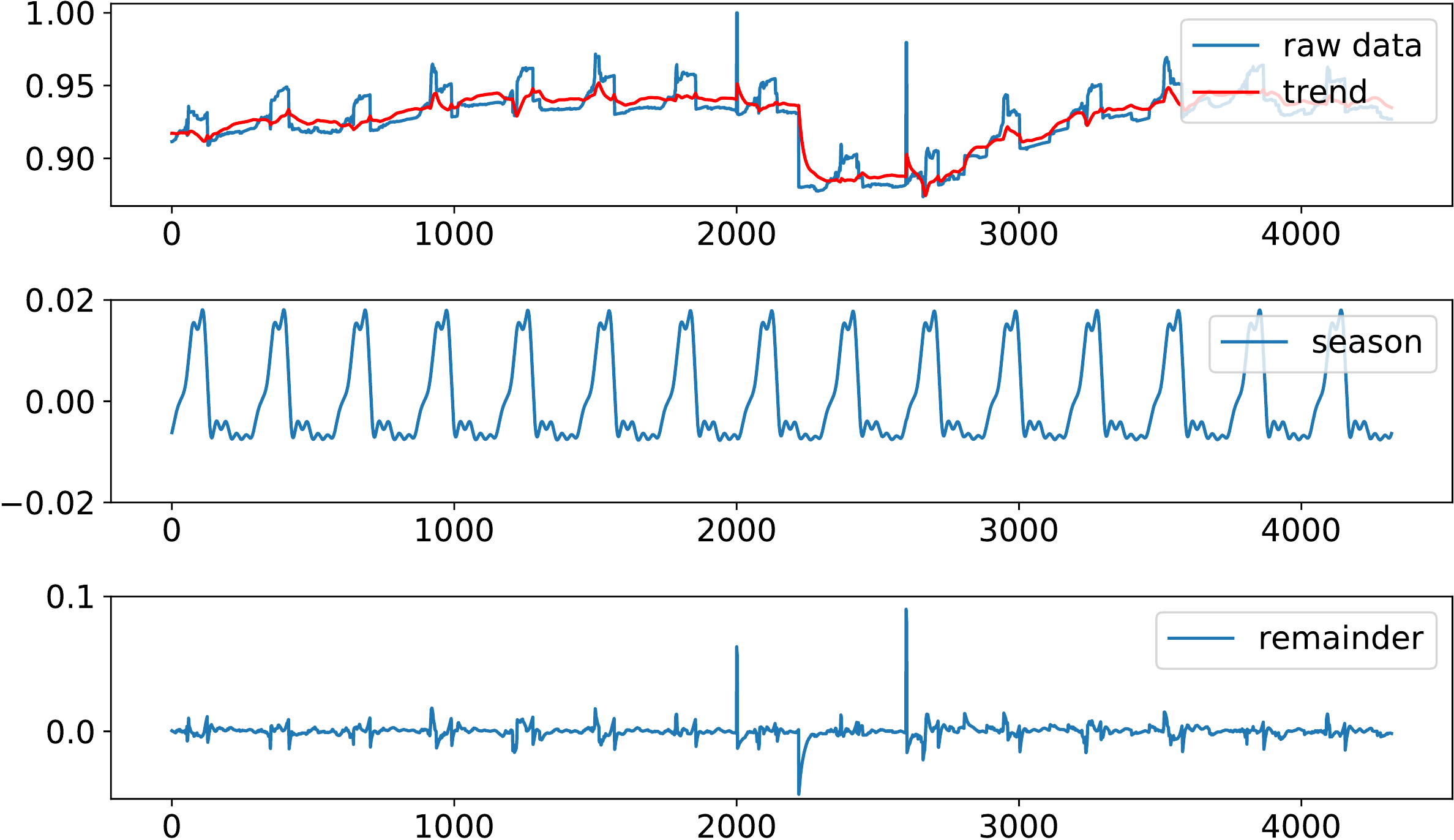}
        }
    \caption{Decomposition results on real dataset 2 for a) Robust STL; b) Standard STL; and c) TBATS.}
    \label{fig:4-tesla-results}
\end{figure*}

\subsection{Experiments on Real-World Data}
\subsubsection{Dataset}
The real-world datasets to be evaluated include two time series. One is the supermarket and grocery stores turnover from 2000 to 2009~\cite{alexandrov2012review}, which has a total of 120 observations with the period $T=12$. We apply the log transform on the data and inject the trend changes and anomalies to demonstrate the robustness (suggested from ~\cite{alexandrov2012review}). The input data can be seen in the top subplot of Figure~\ref{fig:4-grocery-results} (a). We denote it as \textbf{real dataset 1}. Another time series is the file exchange count number in a computer cluster, which has a total of 4032 observations with the period being 288. We apply the linear transform to convert the data to the range of $[0,1]$ for the purpose of data anonymization. The data can be seen in the top subplot of Figure~\ref{fig:4-tesla-results} (a). We denote it as \textbf{real dataset 2}.

\subsubsection{Experimental Settings}
For all four algorithms, period parameter $T=12$ is used for real-world dataset 1, and $T=288$ is used for dataset 2. The rest parameters are optimized using cross-validation for the standard STL, TBATS, and STR. For our proposed RobustSTL algorithm, we set the neighbour window parameters $K=2, H=2$, the regularization coefficients $\lambda_1=1, \lambda_2=0.5$ for real data 1, and $K=2, H=20, \lambda_1=200, \lambda_2=200$ for real data 2. Notice that as STR is not scalable to time series with long seasonality period, we do not report the decomposition result of dataset 2 for STR.

\subsubsection{Decomposition Results}

Figure~\ref{fig:4-grocery-results} and Figure~\ref{fig:4-tesla-results} show the decomposition results on both real-world datasets. RobustSTL typically extracts smooth seasonal signals on real-world data. The seasonal component can adapt to the changing patterns and the seasonality shifts, as seen in Figure~\ref{fig:4-grocery-results} (a). The extracted trend signal recovers the abrupt changes and level shifts promptly and is robust to the existence of anomalies. The spike anomaly is well preserved in the remainder signal. For the standard STL and STR, the trend does not follow the abrupt change and the seasonal component is highly affected by the level shifts and spike anomalies in the original signal. TBATS can decompose the trend signal that follows the abrupt change, however, the trend is affected by the spike anomalies. 

During the experiment, it is also observed that the computation speed of RobustSTL is significantly faster than TBATS and STR (note the result for STR is not available in Figure~\ref{fig:4-tesla-results} due to its long computation time), as the algorithm can be formulated as an optimization problem with $\ell_1$-norm regularization and solved efficiently. While the standard STL seems computational efficient, its sensitivity to anomalies and incapability to capture the trend change and level shifts make it difficult to be used on enormous real-world complex time series data.

Based on the decomposition results from both the synthetic and the real-world datasets, we conclude that the proposed RobustSTL algorithm outperforms existing solutions in terms of handling abrupt and level changes in the trend signal, the irregular seasonality component and the seasonality shifts, the spike \& dip anomalies effectively and efficiently.

\section{Conclusion}

In this paper, we focus on how to decompose complex long time series into trend, seasonality, and remainder components with the capability to handle the existence of anomalies, respond promptly to the abrupt trend changes shifts, deal with seasonality shifts \& fluctuations and noises, and compute efficiently for long time series. We propose a robustSTL algorithm using LAD with $\ell_1$-norm regularizations and non-local seasonal filtering to address the aforementioned challenges. Experimental results on both synthetic data and real-world data have demonstrated the effectiveness and especially the practical usefulness of our algorithm. In the future we will work on how to integrate the seasonal-trend decomposition with anomaly detection directly to provide more robust and accurate detection results on various complicated time series data. 



\bibliographystyle{aaai}

\bibliography{reference}
\end{document}